\documentclass[11pt]{article}
\usepackage[letterpaper,margin=1in]{geometry}
\usepackage{amsmath,amssymb,amsthm,mathtools}
\usepackage{newtxtext,newtxmath}
\usepackage{microtype}
\usepackage{graphicx}
\usepackage{booktabs,tabularx,array,multirow}
\usepackage{enumitem}
\usepackage{xcolor}
\usepackage[round,authoryear]{natbib}
\usepackage{url}
\usepackage[hidelinks]{hyperref}
\usepackage{caption}
\usepackage{float}
\newtheorem{assumption}{Assumption}
\newtheorem{proposition}{Proposition}
\newtheorem{lemma}{Lemma}
\newtheorem{theorem}{Theorem}
\newcommand{\E}{\mathbb{E}}
\newcommand{\Prob}{\mathbb{P}}
\newcommand{\cE}{\mathcal{E}}
\newcommand{\cF}{\mathcal{F}}
\newcommand{\cG}{\mathcal{G}}
\newcommand{\best}[1]{\textbf{#1}}
\title{\textbf{Evidence-Gated Research: Statistically Controlled Model Adoption in Adaptive Search}}
\author{Yifan Guo\\Department of Applied Mathematics and Statistics\\Stony Brook University}
\date{Preprint -- October 2026}
\begin{document}
\maketitle

\begin{abstract}
Adaptive model search is path dependent: once a challenger is adopted, it becomes the reference from which later candidates are generated. A statistically unsupported replacement can therefore alter hypotheses that have not yet been proposed. We introduce Evidence-Gated Research (EGR), a statistical adoption layer for moving-incumbent search. EGR freezes each challenger before decision evidence is revealed, builds anytime-valid evidence across a predeclared set of environments, routes evidence predictably toward unresolved components, composes a persistent candidate e-value, and passes that e-value to an online controller. Under explicit conditional-validity and predictability conditions, the resulting procedure controls false discovery rate for the declared all-environment adoption target even though earlier adoptions change later challengers. In a 5,000-trajectory closed-loop benchmark, development-only e-LOND attains persistent FDR 0.621, whereas no persistent false-adoption path is observed for the audited EGR variants in that finite run. In matched replay over 600 challenger--incumbent pairs, Stagewise EGR preserves fixed-anytime alternative crossing decisions while using 56.1\% less decision evidence at the representative threshold. A three-environment public-data study and a 40,000-sample controlled neural benchmark reproduce the evidence-efficiency pattern. These results identify model replacement as a distinct statistical control point in adaptive model development.
\end{abstract}

\section{Introduction}
Modern model development is increasingly closed-loop: propose a modification, evaluate it, decide whether to replace the incumbent, and generate the next proposal from the resulting state. AutoML and hyperparameter optimization make this coupling explicit \citep{baratchi2024automl}, while automated research agents increasingly instantiate the same search--evaluate--update loop \citep{chan2025mlebench,toledo2025agents,lu2026aiscientist,aygun2026era,ghareeb2026robin}. Recent systems now use tree search to optimize empirical scientific software and multi-agent loops to generate hypotheses, propose experiments, and analyze results \citep{aygun2026era,ghareeb2026robin}. In such systems, an unsupported replacement is \emph{state changing}: it changes the reference model from which later candidates are generated.

This creates a statistical problem that is not captured by ordinary fixed-family model comparison: \emph{when is there enough evidence to let an adaptively proposed challenger replace the current incumbent?} Four issues interact. A development-selected gain may fail in another environment that is part of the adoption target. Evidence may be stopped or redirected as it arrives, so candidate-level inference must remain valid under sequential acquisition. Repeated challenger tests create stream-level multiplicity even if every individual candidate test is anytime valid. Finally, a fixed acquisition plan can waste observations on environments whose evidence is no longer decision-limiting.

Evidence-Gated Research (EGR) separates flexible proposal generation from the statistical decision to replace an incumbent. At step $t$, a predictable policy proposes challenger $c_t$ against incumbent $b_{t-1}$. EGR fixes that comparison before decision evidence is opened, tests whether the gain exceeds a predeclared practical margin in every declared environment, and permits adoption only when the resulting persistent candidate e-value clears a stream-level threshold. The proposal mechanism may remain fully adaptive across candidates: validity is required conditionally on the realized history that generated the current frozen comparison.

\paragraph{Contributions.}
(1) We formulate model replacement as an online sequence of history-dependent, moving-incumbent hypotheses and define persistent benefit relative to predeclared environments and practical margins. (2) We give sufficient conditions under which predictably routed, anytime-valid component evidence yields a conditionally valid persistent candidate e-value, and show how this candidate interface composes with online e-value control for persistent-target FDR (or FWER under a stricter threshold). (3) We separate validity from evidence acquisition: Stagewise EGR short-circuits resolved intersection--union decisions, EGR-ALLOC recycles saved budget, and a balancing rule targets weak components. Experiments then isolate target mismatch, stream multiplicity, evidence-unit dependence, candidate-time leakage, controller choice, and acquisition cost.

\begin{figure}[t]
\centering
\includegraphics[width=.98\linewidth]{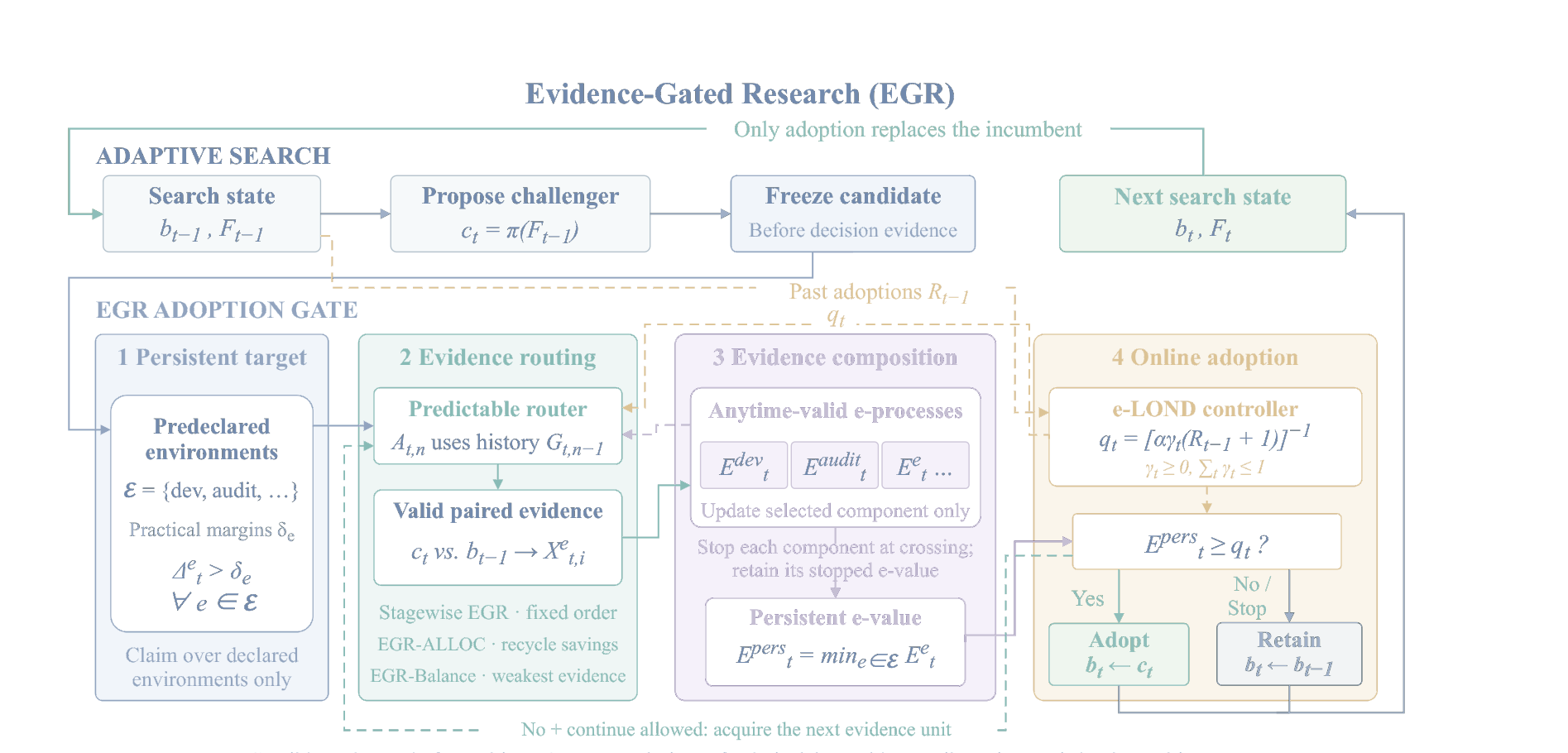}
\caption{EGR inserts a statistically controlled adoption gate between adaptive proposal generation and the next search state. The challenger is frozen before decision evidence is opened; component e-processes are updated under predictable routing, combined through a persistent target, and passed to a compatible online controller. Only adoption changes the incumbent.}
\label{fig:framework}
\end{figure}

\section{Related Work}
\paragraph{Adaptive validation and model search.}
The reusable holdout studies repeated adaptive access to evaluation data \citep{dwork2015holdout}. Learn-then-test (LTT) converts risk calibration into multiple testing \citep{angelopoulos2025ltt}, and adaptive LTT (aLTT) uses e-processes for sequential hyperparameter selection \citep{zecchin2025altt}. Model selection under temporal shift estimates changing-environment performance and compares candidate models \citep{han2024shift}. Recent automated-science systems broaden adaptive search to empirical software optimization and multi-agent discovery workflows \citep{lu2026aiscientist,aygun2026era,ghareeb2026robin}; these works motivate increasingly autonomous proposal loops but do not supply the moving-incumbent, multi-environment adoption control studied here. EGR instead targets a state-changing challenger--incumbent stream with a predeclared intersection target across environments.

\paragraph{Anytime evidence, experimentation, and online multiplicity.}
Always-valid A/B inference permits continuous monitoring under optional stopping \citep{johari2022alwaysvalid}. Betting-based e-processes provide a general anytime-valid framework \citep{ramdas2023savi,waudby2024betting,pandeva2024deep,ramdas2025evalues}. Once a valid candidate e-value is available, e-LOND controls online FDR for e-value streams \citep{xu2024elond}; e-GAI methods such as e-LORD and e-SAFFRON improve spending efficiency \citep{zhang2025egai}, and online e-closure/compound-e methods provide further power improvements \citep{xu2026eclosure}. EGR addresses the upstream construction and adaptive acquisition of the \emph{candidate-level} multi-environment evidence required by such controllers.

\paragraph{Intersection--union targets.}
Persistent benefit has classical intersection--union structure: the scientific alternative holds only if every environment-specific alternative holds \citep{berger1996iut}. EGR combines this structure with anytime-valid components and a changing incumbent.

\begin{table}[t]
\centering\scriptsize
\caption{Statistical objects addressed by related frameworks. Our construction is shown in the final column.}
\resizebox{\linewidth}{!}{%
\begin{tabular}{@{}llll@{}}
\toprule
Object & aLTT & e-LOND / e-GAI & \textbf{EGR (ours)}\\
\midrule
Native target & candidate / hyperparameter risk & online e-value stream & \textbf{persistent challenger--incumbent adoption}\\
Within-candidate acquisition & sequential & assumes e-value supplied & \textbf{predictable multi-env. acquisition}\\
Changing incumbent & extra specification & extra specification & \textbf{explicitly modeled}\\
Across-candidate control & family selection & online FDR & \textbf{compatible online controller}\\
\bottomrule
\end{tabular}}
\label{tab:positioning}
\end{table}

\section{Persistent Adoption in an Adaptive Search Loop}
Let $\cF_{t-1}$ contain all proposals, measurements, and adoption decisions available before candidate $t$. A predictable proposal policy selects
\begin{equation}
 c_t=\pi(\cF_{t-1}),
\end{equation}
and $c_t$ is compared with current incumbent $b_{t-1}$. For each predeclared environment $e\in\cE$, define paired utility gain
\begin{equation}
 X^e_{t,i}=U^e(c_t;Z^e_{t,i})-U^e(b_{t-1};Z^e_{t,i}),\qquad
 \Delta_t^e=\E[X^e_{t,i}\mid\cF_{t-1}],
\end{equation}
where the practical margin $\delta_e\ge0$ is fixed before candidate-$t$ decision evidence is revealed. We call $c_t$ \emph{persistently beneficial} when $\Delta_t^e>\delta_e$ for every $e\in\cE$. The persistent null is therefore
\begin{equation}
 H_t^{\mathrm{pers}}=\bigcup_{e\in\cE}H_t^e,\qquad H_t^e:\Delta_t^e\le\delta_e.
\end{equation}
An adoption made while $H_t^{\mathrm{pers}}$ is true is a \emph{persistent false adoption}. If $R_T$ and $V_T$ are total and false adoptions through $T$, we target
\begin{equation}
 \mathrm{FDR}(T)=\E\!\left[\frac{V_T}{R_T\vee1}\right],\qquad
 \mathrm{FWER}(T)=\Prob(V_T\ge1).
\end{equation}
``Persistent'' is local to the declared all-environment target; it does not imply robustness to undeclared future shifts.

\section{Evidence-Gated Research}
\subsection{Anytime-valid components and persistent composition}
For candidate $t$, let $\cG_{t,n-1}$ denote the full joint history before global allocation $n$, let $A_{t,n}\in\cE$ be the environment selected next, and let $N^e_{t,n}=\sum_{s\le n}\mathbf 1\{A_{t,s}=e\}$ count the number of environment-$e$ updates revealed by that time.

\begin{assumption}[Sequentially valid evidence unit]
Whenever $H_t^e$ is true, the predeclared acquisition mechanism yields bounded selected updates satisfying
\begin{equation}
\E\!\left[X^e_{t,N^e_{t,n}}-\delta_e\mid\cG_{t,n-1},A_{t,n}=e\right]\le0.
\label{eq:unit}
\end{equation}
\end{assumption}
The condition is deliberately joint-history conditional: it must hold for the actual unit consumed by the sequential update. Fresh independent draws satisfy it directly; with dependence, the valid unit may instead be a block or cluster. A marginal mean condition for correlated raw rows is not sufficient.

Choose $B_e$ so that $Y^e_{t,i}=(X^e_{t,i}-\delta_e)/B_e\in[-1,1]$. For precommitted $0\le\lambda_k<1$, define
\begin{equation}
E^e_{t,m}=\frac1K\sum_{k=1}^K\prod_{i=1}^m(1+\lambda_kY^e_{t,i}).
\label{eq:eprocess}
\end{equation}
\begin{proposition}[Component e-process]
Under Assumption~1, each betting product in Eq.~\eqref{eq:eprocess} is a nonnegative supermartingale on its component-null event. Their fixed mixture is therefore an e-process. At any almost-surely finite stopping time $\tau$,
\[
\E[\nu_t^eE^e_{t,\tau}\mid\cF_{t-1}]\le\nu_t^e,
\]
where $\nu_t^e=\mathbf 1\{H_t^e\text{ is true}\}$.
\end{proposition}

\begin{lemma}[Persistent candidate e-value]\label{lem:min}
Suppose each nonnegative stopped component value satisfies
$\E[\nu_t^eE_t^e\mid\cF_{t-1}]\le\nu_t^e$.
Then
\begin{equation}
E_t^{\mathrm{pers}}=\min_{e\in\cE} E_t^e
\label{eq:min}
\end{equation}
satisfies $\E[\nu_tE_t^{\mathrm{pers}}\mid\cF_{t-1}]\le\nu_t$, where $\nu_t=\max_e\nu_t^e$. No independence across environments is required.
\end{lemma}
Under the union null, at least one component is null and Eq.~\eqref{eq:min} is pointwise bounded by that component. The price of this assumption-light composition is possible conservatism when one environment accumulates evidence slowly.

\subsection{Predictable routing and evidence allocation}
\begin{theorem}[Validity under predictable routing]\label{thm:routing}
If $A_{t,n}$ is $\cG_{t,n-1}$-measurable and Assumption~1 holds, then updating a component only when its environment is selected preserves its nonnegative-supermartingale property on the component-null event. Any almost-surely finite joint-history stopping rule therefore produces a conditionally valid component e-value, and the stopped components may be combined by Lemma~\ref{lem:min}.
\end{theorem}

We use three routing rules. \textbf{Stagewise EGR} visits environments in a fixed predeclared order and opens the next component only after the current component crosses its candidate threshold. \textbf{EGR-ALLOC} has nominal cap $B$ per environment and total cap $|\cE|B$; at stage $j$ (zero indexed), cumulative spend may not exceed $(j+1)B$, so the current stage can recycle only budget saved by completed stages while reserving $B$ for every unopened future stage. \textbf{Evidence balancing} sends the next unit to an unresolved environment with the weakest current log evidence.

For intuition, suppose deterministic log evidence grows as $\ell_e(n_e)=n_er_e$, $r_e>0$, and define $S=\sum_e r_e^{-1}$ and $r_{\max}=\max_e r_e$.
\begin{proposition}[First-order balancing guarantee]\label{prop:balance}
If each observation is sent to a current minimizer of $n_er_e$, then after $N$ allocations,
\begin{equation}
\frac{N}{S}-r_{\max}\le\min_e n_er_e\le\frac{N}{S}.
\end{equation}
Thus any rule that raises every component to level $L$ needs at least $SL$ observations, while greedy balancing reaches the level within $\lceil S(L+r_{\max})\rceil$. This is a deterministic benchmark, not a general stochastic optimality claim.
\end{proposition}

\begin{table}[t]
\centering\scriptsize
\caption{Operational validity contract. The guarantee is local to the declared target and evidence unit.}
\begin{tabular}{@{}lll@{}}
\toprule
Condition & Required? & Role\\
\midrule
Candidate frozen before evidence & Yes & prevents candidate-time leakage\\
Joint-filtration validity & Yes & links null to selected updates\\
Predictable allocation / stopping & Yes & preserves e-process validity\\
Cross-environment independence & No & minimum composition is pointwise\\
Raw-unit dependence & Not automatic & must be handled in evidence unit\\
\bottomrule
\end{tabular}
\label{tab:contract}
\end{table}

\subsection{Across-candidate control}
For nonnegative $\gamma_t$ with $\sum_t\gamma_t\le1$, e-LOND uses \citep{xu2024elond}
\begin{equation}
\alpha_t=\alpha\gamma_t(R_{t-1}+1),\qquad q_t=\alpha_t^{-1},
\end{equation}
and adopts only if $E_t^{\mathrm{pers}}\ge q_t$. Other compatible online e-value controllers may be substituted.

\begin{theorem}[End-to-end persistent adoption control]\label{thm:main}
Assume (i) $c_t$ is fixed before candidate-$t$ decision evidence is opened, (ii) Assumption~1 holds for every true component null, and (iii) allocation and stopping are predictable. Then $E_t^{\mathrm{pers}}$ is conditionally valid under every true persistent null given $\cF_{t-1}$. Feeding these e-values to e-LOND yields online persistent-target FDR control at level $\alpha$ under the dependence conditions of \citet{xu2024elond}, even though earlier adoptions may affect future proposals. Replacing the threshold by $1/(\alpha\gamma_t)$ yields persistent FWER at most $\alpha$.
\end{theorem}
Complete proofs are in Appendix~\ref{app:proofs}. Scout data used to construct a challenger must remain separate from its decision evidence; any post-evidence change to architecture, prompt, hyperparameters, weights, or other decision-relevant attributes defines a new candidate and restarts both its evidence process and downstream testing index.

\section{Experiments}
Our experiments ask: \textbf{Q1} why development-only validation can fail the persistent target; \textbf{Q2} how repeated state-changing adoptions interact with stream multiplicity; and \textbf{Q3} whether predictable evidence placement reduces validation cost without changing the declared target. Finite-run error estimates below are descriptive; formal guarantees are Theorem~\ref{thm:main}. Full protocols, uncertainty estimates, and additional diagnostics are in Appendices~\ref{app:protocols}--\ref{app:diagnostics}.

\subsection{Target mismatch and repeated adoption}
The headline synthetic closed loop has two search coordinates: one improves both environments and the other is rewarded on development but penalized on audit. Each of $T=20$ steps scouts $M$ mutations of the current incumbent; accepted challengers become parents of later proposals. At $M=16$, 5,000 trajectories, $\delta=0.01$, and per-environment reference cap $B=120$, development-only e-LOND remains valid for the development hypothesis yet attains persistent FDR $0.621$. Reusing an anytime-valid two-environment candidate test without stream correction yields both all-null FDR and $\Prob(V_T>0)$ of $0.999$ by $T=500$, demonstrating that candidate-level anytime validity does not by itself control a repeated adoption stream.

\begin{figure}[t]
\centering
\includegraphics[width=.80\linewidth]{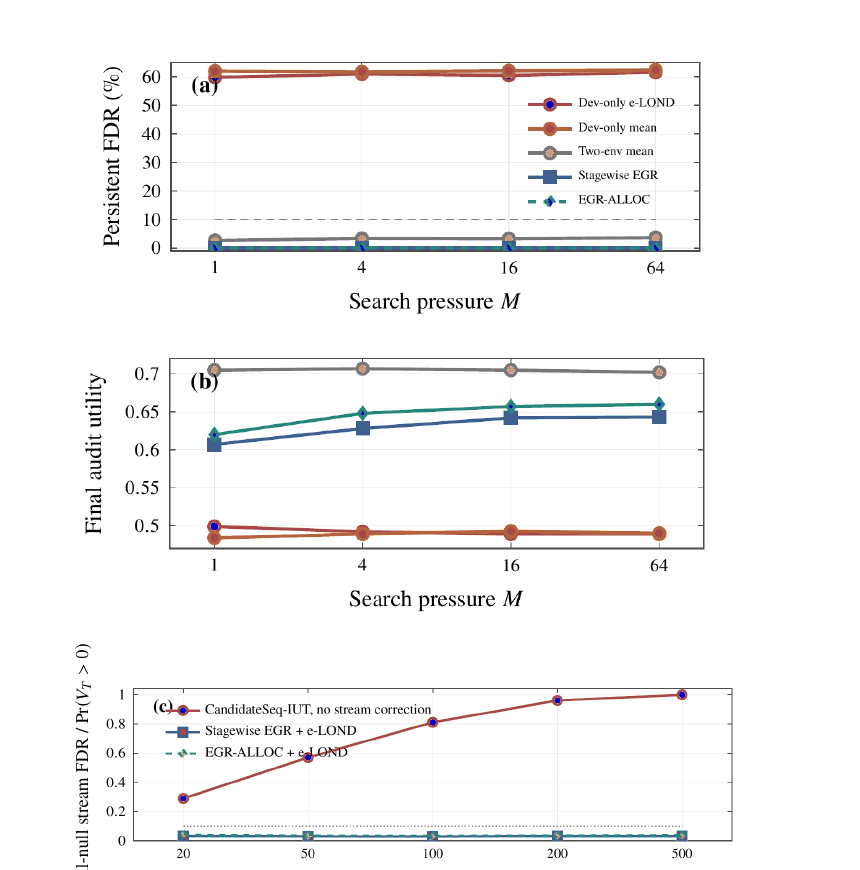}
\caption{Target mismatch and stream multiplicity. Development-only procedures can appear powerful while missing the declared persistent target; repeated reuse of an anytime-valid candidate test accumulates false adoptions, whereas EGR with online correction remains controlled in the tested regimes.}
\label{fig:closedloop}
\end{figure}

\begin{table}[t]
\centering\scriptsize
\setlength{\tabcolsep}{2.7pt}
\caption{Headline closed-loop results. Methods are columns so EGR variants appear last. Lower is better for FDR/evidence; higher is better for power/audit utility.}
\begin{tabular}{@{}lrrrrrr@{}}
\toprule
Metric & \shortstack{Dev\\mean} & \shortstack{Dev\\e-LOND} & \shortstack{Two-env\\mean} & \shortstack{Fixed IUT\\+ e-LOND} & \shortstack{\textbf{Stagewise}\\\textbf{EGR}} & \textbf{EGR-ALLOC}\\
\midrule
Persistent FDR & .6240 & .6210 & .0340 & .0000 & \best{.0000} & \best{.0000}\\
Pers.-alt. power & .907 & .461 & .867 & .188 & .196 & \best{.251}\\
Final audit utility & .493 & .489 & .705 & .641 & .642 & \best{.657}\\
Evidence / candidate & 240.0 & 221.6 & 240.0 & 240.0 & \best{145.5} & 153.7\\
\bottomrule
\end{tabular}
\label{tab:headline}
\end{table}
Relative to fixed audited IUT, EGR-ALLOC uses 36\% less decision evidence and raises persistent-alternative power from .188 to .251. Neither EGR variant has an observed persistent false-adoption path in the 5,000 headline trajectories; $0/5000$ corresponds to a one-sided 95\% binomial upper bound of approximately $6\times10^{-4}$.

\subsection{Stream-level controller sensitivity}
To separate candidate construction from downstream spending, we feed identical persistent candidate e-value streams to four compatible controllers at $T=500$ with 5\% alternatives. Persistent-alternative power is .619 for e-LOND, .890 for e-LORD, .888 for e-SAFFRON, and .415 for the e-FWER rule; the corresponding empirical FDRs are .00015, .00055, .00055, and .00020. Thus controller choice materially affects long-horizon power even when the EGR candidate interface is held fixed.

\subsection{Evidence validity, candidate freezing, and adaptive allocation}
\paragraph{Evidence-unit dependence.}
A clustered-observation stress test probes Assumption~1. At within-cluster correlation $\rho=.95$, treating dependent raw rows as fresh sequential evidence units crosses the candidate null threshold with probability .728, whereas aggregating each cluster into one bounded evidence unit reduces the crossing probability to .0127. This directly shows that marginal mean validity is insufficient when the sequential evidence unit is misspecified.

\paragraph{Candidate-time leakage.}
We next stress-test the requirement that a challenger be frozen before decision evidence is opened. Using the same synthetic utility surface and paired Bernoulli evidence mechanism, we construct development-favorable challengers exactly on the audit boundary, so every variant is a true persistent null. To isolate leakage from optional stopping, component e-values are evaluated at a fixed endpoint. At candidate threshold $q=20$ with $L=64$ post-evidence variants, a challenger frozen using scout data crosses the persistent null in only .052\% of 100,000 trials. If the decision evidence is instead used to select the most favorable variant and then reused for adoption, the crossing rate rises to 3.802\% (about $73\times$ the frozen rate). Making the same post-evidence selection but discarding those observations and restarting on fresh decision evidence restores the rate to .063\%. The failure therefore comes from reusing candidate-shaping evidence, not from adaptivity itself. Full protocol and threshold sweeps appear in Appendices~\ref{app:protocols}--\ref{app:diagnostics}.

\paragraph{Adaptive allocation.}
Once the target and evidence unit are valid, allocation is an efficiency problem. Across $|\cE|\in\{2,3,5\}$ and $q\in\{5,10,20,50,100\}$, weakest-component balancing uses only realized evidence yet stays close to an oracle supplied with simulator evidence rates: the balancing/oracle mean-evidence ratio remains between .970 and 1.034 in the tested regimes.

\subsection{Multi-environment and larger neural searches}
The original public-data experiments use a fixed stratified scikit-learn digits snapshot, local incumbent-changing proposals, a held-out hidden environment used only for evaluation, and paired correctness gains sampled with replacement from fixed empirical decision pools. We extend this setting from two to \textbf{three declared environments}: clean development, Gaussian-noise audit, and dropout+noise audit, with a stronger corruption reserved for hidden evaluation (E30). The sequential statement is therefore relative to the induced finite empirical distributions, not to an unknown deployment population.

We additionally run a larger controlled neural benchmark (E32) with 40,000 samples, 50 features, five classes, a 16-model MLP bank, three declared environments, and a stronger evaluation-only hidden environment. This experiment tests scale/mechanism robustness while remaining controlled/synthetic; it is not a CIFAR/ImageNet or LLM-agent benchmark.

\begin{table}[t]
\centering\footnotesize
\setlength{\tabcolsep}{6pt}
\caption{Three-environment audited comparisons. EGR columns are last; Monte Carlo uncertainty is reported in Appendix~\ref{app:diagnostics}.}
\begin{tabular}{@{}llrrr@{}}
\toprule
Benchmark & Metric & \shortstack{Fixed IUT\\+ e-LOND} & \textbf{Stagewise EGR} & \textbf{EGR-ALLOC}\\
\midrule
Digits 3-env. & FDR & 0 & \best{0} & \best{0}\\
($N=800$) & Power & .3137 & .3221 & \best{.3349}\\
 & Evidence & 1200.0 & \best{388.8} & 392.6\\
\addlinespace
Controlled 40k & FDR & 0 & \best{0} & \best{0}\\
($N=500$) & Power & .2149 & .2380 & \best{.2666}\\
 & Evidence & 1200.0 & 444.6 & \best{440.9}\\
\bottomrule
\end{tabular}
\label{tab:multienv}
\end{table}
Stagewise EGR reduces mean evidence by 67.6\% versus fixed IUT in the three-environment digits extension and by 62.9\% in the 40k controlled benchmark. EGR-ALLOC preserves the same total cap while recycling only saved earlier-stage budget; it improves power to .3349 and .2666, respectively, with no observed persistent false-adoption path in those audited runs. Proposal-policy robustness experiments (local, wide-local, global predictable proposals) are reported in Appendix~\ref{app:diagnostics}.

\begin{figure}[t]
\centering
\includegraphics[width=.78\linewidth]{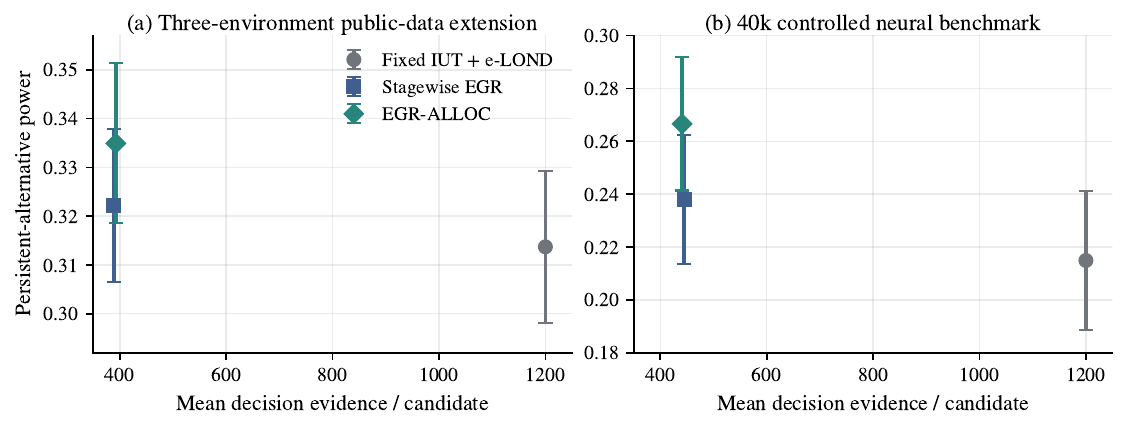}
\caption{Persistent-alternative power versus decision-evidence cost in the three-environment studies. Error bars show 95\% Monte Carlo intervals. The EGR variants achieve substantially lower evidence use; exact numerical summaries are in Table~\ref{tab:multienv}.}
\label{fig:multienv}
\end{figure}

\subsection{Matched replay isolates acquisition efficiency}
To remove proposal-state feedback, we freeze all 600 ordered non-self challenger--incumbent pairs from the 25-model neural grid and replay identical paired evidence streams across acquisition rules. At $q=20$, fixed-anytime IUT and Stagewise EGR have identical component crossing events by construction, hence identical persistent-alternative power (.6477), while Stagewise reduces mean evidence from 1600.0 to 702.4 (56.1\%). The paired saving is 897.6 observations per candidate (95\% pair-bootstrap CI 878.1--917.3). EGR-ALLOC reaches power .6699 at 715.7 observations; its power increase over Stagewise is .0222 (95\% CI .0159--.0290).

\begin{figure}[t]
\centering
\includegraphics[width=.78\linewidth]{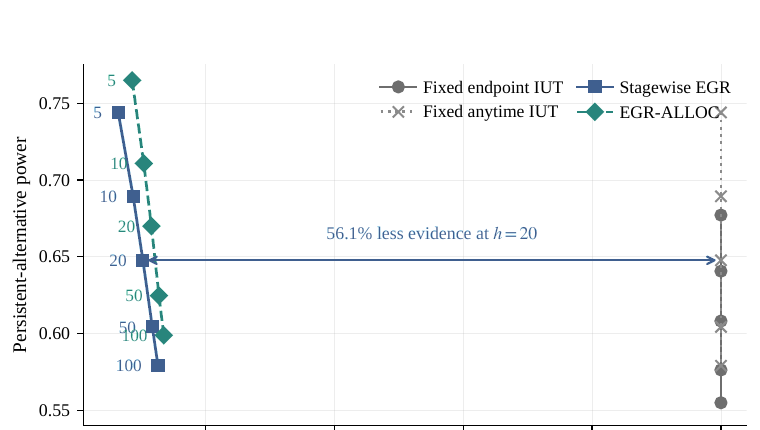}
\caption{Matched frozen-candidate replay over 600 challenger--incumbent pairs. Stagewise EGR has the same alternative crossing event as fixed-anytime IUT while placing evidence sequentially; EGR-ALLOC can recycle saved early-stage budget. At $q=20$, Stagewise uses 56.1\% less evidence at identical power.}
\label{fig:frozen}
\end{figure}

\section{Discussion and Limitations}
The experiments separate five failure modes that are easy to conflate. Target mismatch is not repaired by online multiplicity correction: a development-only procedure can be statistically valid for the wrong scientific target. Candidate-level anytime validity does not remove across-candidate multiplicity. Valid component evidence can fail when dependent raw rows are treated as fresh sequential units. Candidate-level validity can also fail when decision evidence is allowed to shape the candidate and then reused: the leakage stress test increases a $q=20$ boundary-null crossing rate from .052\% to 3.802\%, while a fresh restart restores .063\%. Finally, once target, evidence unit, candidate identity, and stream correction are fixed, evidence placement is an efficiency problem: the matched replay isolates a large reduction in acquisition cost without changing Stagewise crossing decisions.

The framework remains deliberately modular. Controller choice affects long-horizon power (Appendix~\ref{app:diagnostics}); the candidate layer can be paired with stronger compatible online e-value controllers as their assumptions permit. Guarantees are conditional on the declared utility, environments, margins, valid evidence unit, and candidate-freezing/restart protocol; they do not imply robustness to undeclared shifts. The leakage experiment is a mechanism stress test with boundary-null challengers and independent evidence streams across candidate variants; it is not a deployment-risk estimate. The public-data studies use finite empirical pools, and the 40k neural benchmark is controlled/synthetic. Larger prospectively sampled image, language-model, and fully agentic search loops remain important tests of external scope. Runtime overhead in large training systems is also not characterized here; our experiments measure statistical evidence cost.

\section{Conclusion}
EGR treats state-changing model adoption as a separate statistical control point in adaptive search. By freezing challengers, constructing predictably routed multi-environment anytime evidence, composing a persistent candidate e-value, and applying stream-level online control, the framework separates proposal flexibility from adoption validity. Across closed-loop, leakage, public-data, matched-replay, multi-environment, and larger controlled-neural studies, the empirical pattern is that target mismatch, candidate-time leakage, or missing stream correction can drive unsupported adoptions, while predictable evidence placement can substantially reduce validation cost once the target and validity contract are correctly specified.

\paragraph{Reproducibility Statement.}
Appendix~\ref{app:protocols} specifies the evidence unit, splits, model families, search policies, budgets, and Monte Carlo conventions. Appendix~\ref{app:proofs} contains complete proofs. The supplementary experiment bundle includes runnable code, source arrays/data, run manifests, paper-facing tables/figures, Monte Carlo uncertainty outputs, reproducibility checks, and SHA-256 checksums for E27/E29 and the E30--E33 experiments.

\paragraph{AI Use Statement.}
Generative AI tools supported parts of the research workflow, including refinement of hypotheses and experiments, analysis-code assistance, mathematical proof drafting, literature search, figure preparation, interpretation of results, and manuscript editing. The author checked the resulting arguments, numerical claims, references, code, and released artifacts against executable analyses and source material and takes responsibility for the final paper.

\appendix
\section{Proofs}\label{app:proofs}

We use the same filtration throughout the appendix. For candidate $t$, $\cF_{t-1}$ is the frozen pre-candidate history and $\cG_{t,n}$ is the joint within-candidate history after $n$ routed allocations. The challenger, incumbent, declared environment laws, utilities, and practical margins are $\cF_{t-1}$-measurable. Let
\[
\nu_t^e=\mathbf 1\{H_t^e\text{ is true}\},\qquad
\nu_t=\max_{e\in\cE}\nu_t^e.
\]
Thus $\nu_t=1$ exactly on the persistent null. Writing validity with these indicators avoids treating the set of true component nulls as deterministic.

\subsection{Proof of Proposition 1}
Fix $t$, $e$, and a betting fraction $\lambda_k$. For the local environment-$e$ clock, define
\[
L^{e,k}_{t,m}=\prod_{i=1}^{m}(1+\lambda_kY^e_{t,i}),\qquad L^{e,k}_{t,0}=1.
\]
Because $Y^e_{t,i}\in[-1,1]$ and $0\le\lambda_k<1$, every multiplier is nonnegative. Under the component null, Assumption~1 implies
\[
\E[\nu_t^eY^e_{t,m}\mid \cG_{t,m-1}]\le 0.
\]
Since $\nu_t^e$ is measurable at the frozen history,
\begin{align*}
\E[\nu_t^eL^{e,k}_{t,m}\mid \cG_{t,m-1}]
&=\nu_t^eL^{e,k}_{t,m-1}
 +\lambda_kL^{e,k}_{t,m-1}\E[\nu_t^eY^e_{t,m}\mid \cG_{t,m-1}]\\
&\le \nu_t^eL^{e,k}_{t,m-1}.
\end{align*}
Hence $(\nu_t^eL^{e,k}_{t,m})_m$ is a nonnegative supermartingale. Averaging over the precommitted betting grid preserves nonnegativity and the supermartingale property, so the mixture in Eq.~\eqref{eq:eprocess} is a component e-process.

Let $\tau$ be any almost-surely finite stopping time. For each deterministic $b$, bounded optional sampling gives
\[
\E[\nu_t^eE^e_{t,\tau\wedge b}\mid \cF_{t-1}]\le \nu_t^e.
\]
As $b\to\infty$, nonnegativity and conditional Fatou's lemma yield
\[
\E[\nu_t^eE^e_{t,\tau}\mid \cF_{t-1}]\le \nu_t^e.
\]
Thus the stopped component value is conditionally valid. A finite evidence cap is a simple implementation-level condition ensuring almost-sure finiteness.

\subsection{Proof of Lemma~\ref{lem:min}}
Order the finite environment set as $e_1,\ldots,e_d$ before decision sampling. Define the disjoint, $\cF_{t-1}$-measurable events
\[
B_j=\{\nu_t^{e_j}=1\}\cap\bigcap_{i<j}\{\nu_t^{e_i}=0\},\qquad j=1,\ldots,d.
\]
The events $B_j$ partition $\{\nu_t=1\}$ and select the first true component null without using decision evidence. Pointwise,
\[
\nu_tE_t^{\mathrm{pers}}
=\nu_t\min_{e\in\cE}E_t^e
\le \sum_{j=1}^{d}\mathbf 1_{B_j}E_t^{e_j}.
\]
Taking conditional expectations and using $\mathbf 1_{B_j}\le \nu_t^{e_j}$ gives
\begin{align*}
\E[\nu_tE_t^{\mathrm{pers}}\mid\cF_{t-1}]
&\le \sum_{j=1}^{d}\mathbf 1_{B_j}
  \E[\nu_t^{e_j}E_t^{e_j}\mid\cF_{t-1}]\\
&\le \sum_{j=1}^{d}\mathbf 1_{B_j}=\nu_t.
\end{align*}
Therefore $E_t^{\mathrm{pers}}$ is a conditional e-value for the union null. The argument is pointwise and requires neither cross-environment independence nor simultaneous component stopping.

\subsection{Proof of Theorem~\ref{thm:routing}}
Fix $t$ and $e$. Let $N^e_{t,n}=\sum_{s\le n}\mathbf 1\{A_{t,s}=e\}$ be the number of environment-$e$ updates by global allocation time $n$. Define the routed increment
\[
W^e_{t,n}=
\begin{cases}
(X^e_{t,N^e_{t,n}}-\delta_e)/B_e, & A_{t,n}=e,\\
0, & A_{t,n}\neq e.
\end{cases}
\]
Because $A_{t,n}$ is $\cG_{t,n-1}$-measurable, the routing decision is known before the next outcome is revealed. Assumption~1 therefore implies
\[
\E[\nu_t^eW^e_{t,n}\mid\cG_{t,n-1}]\le0.
\]
For a fixed $\lambda_k$, define the global-clock product
\[
\widetilde L^{e,k}_{t,n}=\prod_{s=1}^{n}(1+\lambda_kW^e_{t,s}).
\]
Zero increments contribute a factor of one, so this product is exactly the local environment-$e$ betting process evaluated at $N^e_{t,n}$. Moreover,
\begin{align*}
\E[\nu_t^e\widetilde L^{e,k}_{t,n}\mid\cG_{t,n-1}]
&=\nu_t^e\widetilde L^{e,k}_{t,n-1}
 +\lambda_k\widetilde L^{e,k}_{t,n-1}
   \E[\nu_t^eW^e_{t,n}\mid\cG_{t,n-1}]\\
&\le \nu_t^e\widetilde L^{e,k}_{t,n-1}.
\end{align*}
Thus predictable interleaving preserves the component supermartingale property. Averaging over $k$ and applying the stopping argument from Proposition~1 gives a conditionally valid stopped value for every routed component. Lemma~\ref{lem:min} then yields a valid persistent candidate e-value.

\subsection{Proof of Proposition~\ref{prop:balance}}
Let $p_e=n_er_e$ denote deterministic progress after $N$ allocations and let $m_N=\min_ep_e$. Since $p_e\ge m_N$,
\[
N=\sum_en_e=\sum_e\frac{p_e}{r_e}
\ge m_N\sum_e r_e^{-1}=m_NS,
\]
so $m_N\le N/S$.

For the reverse direction, consider any environment $e$ that receives at least one allocation. Immediately before its most recent allocation, greedy balancing selected a current minimizer; after that one update, its progress can exceed the final minimum by at most its own increment size. Hence
\[
p_e\le m_N+r_e\le m_N+r_{\max}.
\]
The same bound is trivial for an environment that is never selected. Therefore
\[
N=\sum_e\frac{p_e}{r_e}
\le m_NS+r_{\max}S,
\]
which gives $m_N\ge N/S-r_{\max}$. To reach level $L$ in every component, any allocation requires $N\ge SL$; greedy balancing reaches the level once $N\ge S(L+r_{\max})$, giving the stated ceiling bound.

\subsection{Proof of Theorem~\ref{thm:main}}
The proof has two layers: candidate-level conditional validity and across-candidate online control.

\paragraph{Candidate-level validity.}
The candidate $c_t$, incumbent $b_{t-1}$, declared target, and all proposal randomness are fixed in $\cF_{t-1}$ before decision evidence is opened. By Proposition~1 and Theorem~\ref{thm:routing}, each true component null yields a stopped value satisfying
\[
\E[\nu_t^eE_t^e\mid\cF_{t-1}]\le\nu_t^e.
\]
Lemma~\ref{lem:min} therefore gives
\[
\E[\nu_tE_t^{\mathrm{pers}}\mid\cF_{t-1}]\le\nu_t.
\]
This statement is conditional on the realized adaptive search history, so previous adoptions may change the next challenger and even the next null hypothesis without invalidating the current candidate e-value.

\paragraph{Across-candidate FDR control.}
The sequence $E_t^{\mathrm{pers}}$ therefore satisfies the conditional e-value requirement supplied to e-LOND. Applying the e-LOND guarantee of \citet{xu2024elond} with
\[
\alpha_t=\alpha\gamma_t(R_{t-1}+1),\qquad
q_t=\alpha_t^{-1},
\]
yields persistent-target FDR control at level $\alpha$ under the controller's stated dependence conditions.

\paragraph{FWER variant.}
If the adoption threshold is instead $q_t=(\alpha\gamma_t)^{-1}$, then on a true persistent null conditional Markov inequality gives
\[
\Prob\!\left(E_t^{\mathrm{pers}}\ge\frac{1}{\alpha\gamma_t}\middle|\cF_{t-1}\right)
\le \alpha\gamma_t\E[E_t^{\mathrm{pers}}\mid\cF_{t-1},\nu_t=1]
\le \alpha\gamma_t.
\]
Taking expectations and applying the union bound over candidate indices gives
\[
\Prob(V_T\ge1)\le\sum_{t\le T}\alpha\gamma_t\le\alpha.
\]
This proves the FWER claim.

\section{Detailed Experimental Protocol}\label{app:protocols}
\subsection{Common conventions}
Empirical FDR is estimated by computing $\mathrm{FDP}=V_T/(R_T\vee1)$ for each independent trajectory and averaging. ``Any false'' is $\Prob(V_T>0)$. Persistent-alternative power is defined relative to the declared all-environment practical-margin target. Evidence counts paired decision-evidence updates across declared environments; scout data used only for candidate proposal are excluded. Component e-processes use the precommitted betting grid $\Lambda=\{.03,.06,.10,.18,.30,.45,.65,.82\}$. For horizon $T$, e-LOND uses $\gamma_t\propto[t\log^2(t+1)]^{-1}$ normalized over $t=1,\ldots,T$. Monte Carlo standard errors are the across-replicate sample standard deviation divided by the square root of the number of nonmissing replicates.

\subsection{Synthetic closed loop and long streams}
The latent stable score is $s(x)=0.20+0.90x_0-0.20x_0^2-0.06x_1^2$, with development and audit logits $\ell_{\rm dev}(x)=s(x)+\rho x_1$ and $\ell_{\rm audit}(x)=s(x)-\kappa x_1$. Bernoulli utilities use the logistic transform. Mutations have Gaussian standard deviation .55 and are clipped to $[-2,2]^2$; the incumbent starts at $(0,0)$. At each step, $M$ mutations are evaluated with four scout observations each; scout observations are never reused as decision evidence. Decision evidence uses common-uniform paired Bernoulli gains with $\delta=.01$. The main grid is $M\in\{1,4,16,64\}$ with 2,000 trajectories per cell; the $M=16$ headline uses 5,000 trajectories. Long-stream diagnostics use $T\in\{20,50,100,200,500\}$, $B=60$, and 2,000 streams per displayed cell.

\subsection{Controller and clustered-dependence diagnostics}
Controller comparisons feed identical persistent candidate e-value streams to e-LOND, e-LORD, e-SAFFRON, and an e-FWER rule. At $T=500$ and 5\% alternatives, powers are .619, .890, .888, and .415, with empirical FDR .00015, .00055, .00055, and .00020. Clustered-dependence experiments vary within-cluster correlation $\rho\in\{0,.25,.5,.75,.95\}$ over 20,000 null trials per setting; the naive construction treats dependent rows as fresh sequential units, whereas the block-valid construction aggregates each cluster into one bounded evidence unit.

\subsection{E33: candidate-freezing / candidate-time leakage stress test}
To isolate candidate-time leakage from optional stopping, E33 evaluates component e-values at a fixed endpoint with $B=800$. It uses the same synthetic score $s(x)$, development/audit logits, practical margin $\delta=.01$, paired Bernoulli gain mechanism, and precommitted betting grid as the main synthetic study. The incumbent is $x=(0,0)$. For each trial and each candidate-pool size $L\in\{1,4,16,64\}$, we sample $x_1\sim\mathrm{Unif}[.8,1.2]$ and solve for the smaller admissible $x_0$ such that the true audit gain equals exactly $\delta$. The resulting development gains lie in [.2766,.3522], while every challenger remains a true persistent null because $\Delta^{\rm audit}=\delta$.

We compare three protocols over 100,000 trials per $L$ and candidate thresholds $q\in\{5,10,20,50\}$. \textbf{Frozen} chooses one variant using four development-only scout observations and only then opens decision evidence. \textbf{Leak+reuse} opens independent decision-evidence streams for all $L$ variants, selects the variant with the largest fixed-endpoint persistent e-value, and reuses those same observations for the decision. \textbf{Leak+restart} makes the same post-evidence selection but treats the leaked observations as candidate-construction data, discards them, and starts a fresh candidate/test index with new decision evidence. Independent streams across variants are deliberate: the diagnostic is designed to expose candidate-time selection bias, not to estimate a deployment error rate.

\subsection{Adaptive-allocation diagnostics}
Allocation studies vary $|\cE|\in\{2,3,5\}$ (with an additional $|\cE|=8$ calibration), candidate threshold $q\in\{5,10,20,50,100\}$, and compare realized weakest-component balancing with an oracle that knows simulator evidence rates.

\subsection{Public-data adaptive search: original two-environment studies}
The fixed stratified digits split contains 808 training, 269 search, 270 development, 270 audit, and 180 hidden examples. Standardization is fit on training only. The RBF-SVC family has 64 configurations, with $\log C\in\{-4,-3,\dots,3\}$ and $\log\gamma\in\{-6,-5.5,\dots,-2.5\}$. The neural family contains 25 one-hidden-layer ReLU MLPs formed by crossing widths $\{4,8,16,32,64\}$ with train-noise augmentation $\{0,.2,.4,.6,.8\}$. Search uses $T=12$, $M=8$ local grid proposals, and 32 scout examples. The neural run uses 1,000 trajectories, $B=800$, $\delta=.005$, $\alpha=.10$. Decision evidence is sampled with replacement from fixed empirical development/audit pools; hidden data are evaluation only.

\subsection{E30: three-environment public-data extension}
E30 reuses the neural digits architecture but declares three environments: clean development, Gaussian-noise audit, and dropout+noise audit. A stronger corruption is hidden evaluation only. Runs use $N=800$, $T=12$, $M=8$, nominal $B=400$ per environment, $\delta=.005$, and $\alpha=.10$. EGR-ALLOC uses the cumulative pooled-budget rule described in Section~4.2. A weakest-component balancing rule is also evaluated. Because decision evidence is sampled with replacement from fixed empirical pools, validity is relative to those empirical distributions.

\subsection{E31: proposal-policy robustness}
E31 fixes the E30 model/evidence bank and adoption rules while changing only the predictable proposal policy: local, wide-local, or global. Each cell uses $N=400$ trajectories. This isolates empirical sensitivity to proposal dynamics without changing the candidate evidence construction.

\subsection{E32: larger controlled neural benchmark}
E32 uses a reproducibly generated 40,000-sample, 50-feature, five-class controlled classification problem and a 16-model MLP bank indexed by hidden width and train-noise augmentation. Three declared environments are clean development, strong additive-noise audit, and dropout+noise audit; a stronger hidden environment is evaluation only. Runs use $N=500$, $T=15$, $M=8$, $B=400$, $\delta=.0025$, and $\alpha=.10$. E32 is a scale/mechanism stress test, not a public-image or LLM-agent benchmark.

\subsection{Frozen-candidate replay}
All 600 ordered non-self pairs from the 25-model neural grid are frozen. Each pair is evaluated with 200 replicated evidence streams at $q\in\{5,10,20,50,100\}$ and per-environment cap $B=800$. Within each pair/replicate, identical development and audit streams are reused across fixed-endpoint IUT, fixed-anytime IUT, Stagewise EGR, and EGR-ALLOC. Therefore fixed-anytime IUT and Stagewise EGR have identical alternative crossing events by construction and differ only in evidence usage.

\section{Additional Results and Diagnostics}\label{app:diagnostics}
\subsection{Original two-environment public-data results}
\begin{table}[ht]
\centering\small
\setlength{\tabcolsep}{8pt}
\caption{Original public-data adaptive searches. EGR methods are the final columns.}
\begin{tabular}{@{}llrrr@{}}
\toprule
Benchmark & Metric & Fixed IUT & \textbf{Stagewise EGR} & \textbf{EGR-ALLOC}\\
\midrule
RBF-SVC & FDR & .0000 & \best{.0000} & \best{.0000}\\
 & Power & .522 & .554 & \best{.583}\\
 & Evidence & 1200.0 & \best{567.5} & 584.3\\
\addlinespace
Neural MLP & FDR & .0000 & .0005 & \best{.0000}\\
 & Power & .319 & .355 & \best{.374}\\
 & Evidence & 1600.0 & 782.3 & \best{781.0}\\
\bottomrule
\end{tabular}
\end{table}

\subsection{Three-environment uncertainty and balancing}
For E30, fixed IUT, Stagewise EGR, and EGR-ALLOC each have 0 observed any-false paths in 800 trajectories; the corresponding one-sided Wilson upper bound is .00478. Power (95\% MC interval) is .3137 [.2981,.3293], .3221 [.3064,.3379], and .3349 [.3185,.3514]. Evidence intervals are 1200 exactly, 388.8 [387.1,390.4], and 392.6 [390.3,394.9]. Weakest-component EGR-Balance attains power .3687 [.3517,.3858] with evidence 1106.4 [1105.4,1107.5] and one observed false-adoption path ($\Prob(V_T>0)=.00125$); it is included as an allocation diagnostic rather than the main efficiency comparison.

For E32, fixed IUT, Stagewise, and EGR-ALLOC have 0 observed any-false paths in 500 trajectories (Wilson upper bound .00762). Power intervals are .2149 [.1886,.2413], .2380 [.2135,.2625], and .2666 [.2414,.2919]; evidence intervals are 1200 exactly, 444.6 [440.2,448.9], and 440.9 [436.6,445.3].

\subsection{Proposal-policy robustness}
\begin{table}[ht]
\centering\small
\setlength{\tabcolsep}{7pt}
\caption{E31 proposal-policy robustness. Every displayed audited method has zero observed FDR in 400 trajectories per policy; EGR methods are the final columns.}
\begin{tabular}{@{}llrrr@{}}
\toprule
Policy & Metric & Fixed IUT & \textbf{Stagewise EGR} & \textbf{EGR-ALLOC}\\
\midrule
Local & Power & .3192 & .3224 & \best{.3309}\\
 & Evidence & 1200.0 & \best{389.3} & 390.7\\
Wide-local & Power & .5466 & .5519 & \best{.5705}\\
 & Evidence & 1200.0 & \best{379.0} & 379.4\\
Global & Power & \best{.6993} & .6877 & .6855\\
 & Evidence & 1200.0 & 379.1 & \best{378.4}\\
\bottomrule
\end{tabular}
\end{table}
The global-policy row illustrates that adaptive evidence placement is primarily an evidence-efficiency device, not a guarantee of higher finite-run power for every proposal dynamic; its validity claim is the conditional one in Theorem~\ref{thm:main}.

\subsection{Frozen-pair diagnostics}
\begin{table}[H]
\centering\small
\setlength{\tabcolsep}{6pt}
\caption{Frozen-pair replay at $q=20$. Reported null rates are candidate-level crossing frequencies, not online-stream FDR.}
\begin{tabular}{@{}lrrrr@{}}
\toprule
Metric & Fixed endpoint & Fixed anytime & \textbf{Stagewise EGR} & \textbf{EGR-ALLOC}\\
\midrule
Pers.-alt. power & .6082 & .6477 & .6477 & \best{.6699}\\
Persistent-null crossing & .000013 & .000355 & .000355 & .000395\\
Mismatch-null crossing & .000000 & .001667 & .001667 & .001905\\
Evidence & 1600.0 & 1600.0 & \best{702.4} & 715.7\\
\bottomrule
\end{tabular}
\end{table}

\subsection{Controller and evidence-unit diagnostics}
At $T=500$ with 5\% alternatives and identical persistent candidate e-values, e-LOND, e-LORD, e-SAFFRON, and the e-FWER rule attain powers .619, .890, .888, and .415, respectively; empirical FDR values remain small in these runs. This shows that downstream spending materially affects long-horizon power while leaving the EGR candidate interface unchanged. In the dependence stress test, treating highly correlated raw rows as fresh evidence produces a null crossing probability .728 at $\rho=.95$; aggregating each cluster into one bounded unit reduces it to .0127, directly illustrating the operational importance of Assumption~1.

\subsection{Candidate-freezing / leakage diagnostic}
The expanded E33 threshold sweep shows that, with one available variant, the frozen and reuse protocols coincide. As the number of post-evidence variants grows, reuse of the same evidence produces a strong multiplicity-like inflation, while restarting remains near the frozen baseline. At $q=20$ and $L=64$, the rates are .052\% (Frozen), 3.802\% (Leak+reuse), and .063\% (Leak+restart).

\begin{table}[H]
\centering\footnotesize
\setlength{\tabcolsep}{8pt}
\caption{E33 persistent-null crossing at $q=20$ over 100,000 trials per $L$ (percent; Wilson 95\% interval in brackets).}
\begin{tabular}{@{}lrrr@{}}
\toprule
Variants $L$ & Frozen & Leak+reuse & Leak+restart\\
\midrule
1  & .063 [.049,.081] & .063 [.049,.081] & .054 [.041,.070]\\
4  & .060 [.047,.077] & .223 [.196,.254] & .042 [.031,.057]\\
16 & .062 [.048,.079] & .988 [.929,1.051] & .059 [.046,.076]\\
64 & .052 [.040,.068] & \best{3.802 [3.685,3.922]} & .063 [.049,.081]\\
\bottomrule
\end{tabular}
\label{tab:freezeleak}
\end{table}

\paragraph{Additional allocation calibration.}
At $q=20$ and $|\cE|=2,3,5,8$, balancing/oracle powers are .898/.904, .961/.965, .996/.996, and 1.000/1.000; mean evidence is 114.1/114.4, 144.4/142.8, 190.8/191.5, and 260.7/258.0, compared with equal-split caps 240, 360, 600, and 960. These are empirical diagnostics under the simulator, not a stochastic optimality theorem.

\end{document}